\documentclass{arx}
\usepackage{cite}
\usepackage{amsmath,amssymb,amsfonts}
\usepackage{algorithmic}
\usepackage{graphicx}
\usepackage{textcomp}
\usepackage{multirow}
\usepackage{float}
\usepackage{url}
\usepackage{hhline}
\usepackage[ruled, vlined, commentsnumbered, linesnumbered]{algorithm2e}
\usepackage{adjustbox}
\title{TransSleep: Transitioning-aware Attention-based \\Deep Neural Network for Sleep Staging}

\author{
  Jauen Phyo$^1$, Wonjun Ko$^1$, Eunjin Jeon$^1$, and Heung-Il Suk$^{1,2}$, \textit{Senior Member, IEEE}\vspace{.1cm} \\
  $^1$Department of Brain and Cognitive Engineering, Korea University \\
  $^2$Department of Artificial Intelligence, Korea University\\
  \{jaeun11, wjko, eunjinjeon, hisuk\}@korea.ac.kr \\
}

\begin{document}
\maketitle

\begin{abstract}
Sleep staging is essential for sleep assessment and plays a vital role as a health indicator. Many recent studies have devised various machine learning as well as deep learning architectures for sleep staging. However, two key challenges hinder the practical use of these architectures: effectively capturing salient waveforms in sleep signals and correctly classifying confusing stages in transitioning epochs. In this study, we propose a novel deep neural network structure, TransSleep, that captures distinctive local temporal patterns and distinguishes confusing stages using two auxiliary tasks. In particular, TransSleep adopts an attention-based multi-scale feature extractor module to capture salient waveforms; a stage-confusion estimator module with a novel auxiliary task, \emph{epoch-level stage classification}, to estimate confidence scores for identifying confusing stages; and a context encoder module with the other novel auxiliary task, \emph{stage-transition detection}, to represent contextual relationships across neighboring epochs. Results show that TransSleep achieves promising performance in automatic sleep staging. The validity of TransSleep is demonstrated by its state-of-the-art performance on two publicly available datasets, Sleep-EDF and MASS. Furthermore, we performed ablations to analyze our results from different perspectives. Based on our overall results, we believe that TransSleep has immense potential to provide new insights into deep learning-based sleep staging.

\end{abstract}
\section{Introduction}
	\label{sec:introduction}
	Human beings spend approximately one-third of their day in sleep, which helps them relieve their body fatigue and maintain essential biorhythms \cite{carskadon2011monitoring}. Thus, sleep plays a critical role in their overall everyday health. Currently, however, an increasing number of people are suffering from sleep disorders such as insomnia and hypnolepsy \cite{han2012stress}. The prevalence of sleep disorders in the general population has been estimated to be in the range of 22-40\% \cite{veldi2005sleep}. Therefore, sleep health has received widespread research attention from the health informatics community.
	
	Generally, polysomnography (PSG), also called a sleep study, is performed to evaluate sleep health and quality. The data recorded during PSG include the subject’s brain waves, such as an electroencephalogram (EEG), and other signals such as an electrooculogram (EOG), electromyogram (EMG), and electrocardiogram (ECG). The PSG recordings are
	typically segmented into pre-defined epochs of 20 or 30 s and categorized into five stages according to the American Academy of Sleep Medicine (AASM): Wake (W), Rapid Eye Movement (REM), Non-REM1 (N1), Non-REM2 (N2), and Non-REM3 (N3) \cite{berry2012rules}.
	
	\begin{figure}[t]
		\centering
		\includegraphics[width=\linewidth]{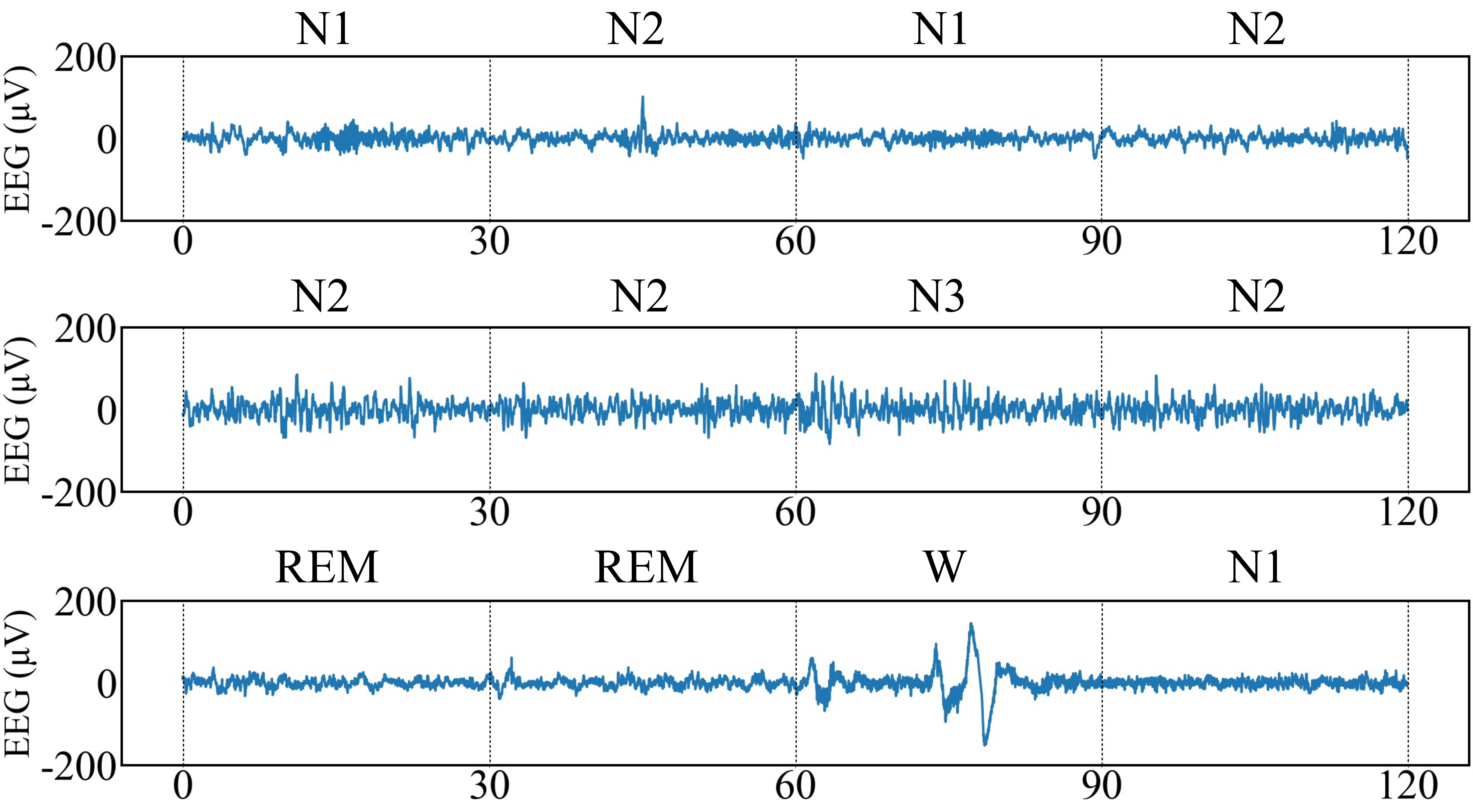}
		\caption{\label{fig1} Transition patterns of sleep stages in the AASM sleep standard. Transition of stages, such as N1-N2-N1-N2, N2-N2-N3-N2, and REM-REM-W-N1, occurs during sleep.}
	\end{figure}
	
	In traditional sleep staging, signals of 8-10 h long sleep stages are manually labeled by human experts according to features of an epoch with transition rules \cite{berry2012rules} that define
	contextual properties among epochs. The brain usually goes through a series of different aspects of signals during sleep, which can be specified using distinct waveform, amplitude, and dominant frequency range (delta, theta, alpha, beta, and gamma waves). For instance, N2 stages are commonly specified by particular patterns such as sleep-spindle and K-complex, whereas a high-amplitude delta rhythm ranging from 1 to 4 Hz is frequently observed at N3 stages \cite{carskadon2011monitoring}. Moreover, several stage-transitioning patterns are also observed, as shown in Fig. 1. For instance, stage transition between adjacent stages in the sleep cycle, such as N1-N2-N1-N2, N2-N2-N3-N2, and REM-REM-W-N1, can be detected \cite{berry2012rules}. Based on these characteristics, human experts manually label sleep stages of 8-10 h long signals; this process, however, is tedious, labor-intensive, and highly subjective.

	Many pioneering studies have proposed and developed various machine learning- or deep learning-based methods for automatic sleep staging \cite{tsinalis2016automatic, alickovic2018ensemble, li2017hyclasss, ghimatgar2019automatic, supratak2017deepsleepnet, qu2020residual, jia2020graphsleepnet, loh2020automated, phan2021xsleepnet}. Some studies \cite{alickovic2018ensemble, li2017hyclasss, ghimatgar2019automatic} have adopted the traditional machine learning approach to classify sleep stages by exploiting hand-engineered features using algorithms such as the support vector machine (SVM) and random forest classifier. Recently, many state-of-the-art deep learning methods \cite{supratak2017deepsleepnet, qu2020residual, jia2020graphsleepnet, phan2021xsleepnet} have utilized a number of linear or non-linear layers without domain knowledge for automatic sleep staging. In particular, these methods use an architectural form of a feature extractor and a context encoder jointly. The feature extractor learns intra-epoch features of an input signal, while the context encoder focuses on representing inter-epoch relations. Hence, the network can extract distinctive features and contextual properties of sleep signals. Nevertheless, these methods suffer from limitations in terms of recognizing salient sleep patterns and confusing stages.

	\begin{figure}[t]
		\centering
		\includegraphics[width=.98\columnwidth]{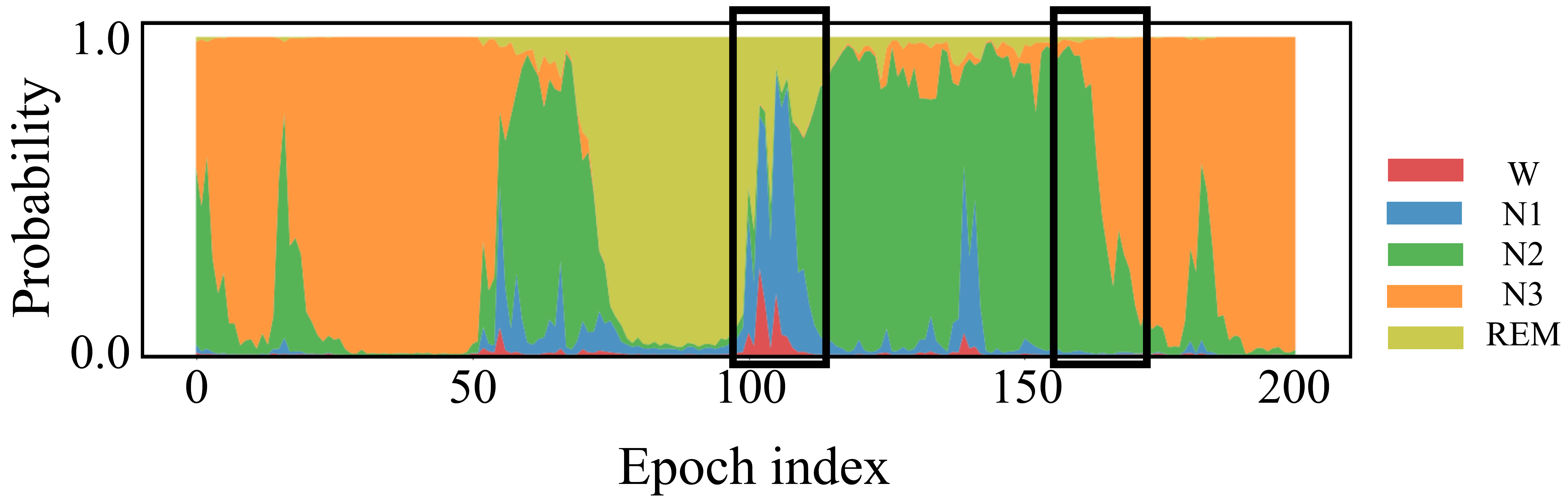}
		\caption{\label{transitioning_epochs} Posterior probability over sleep stages. The solid rectangles denote transitioning epochs, which have multiple stage properties according to sleep stage transition.}
	\end{figure}
	
	For recognizing complex salient patterns, such as alpha rhythm, sleep spindle, K-complex, and slow eye movement, the key is to represent the multi-scale temporal trend in an epoch-level manner \cite{wang2022novel} considering these salient patterns have various spectral and temporal characteristics \cite{aeschbach1993all}. For example, alpha rhythm oscillates periodically with a frequency range of 8 to 12 Hz, whereas K-complex consists of a brief negative high-voltage peak followed by a slower positive complex and occurs approximately every 1.0 to 1.7 min \cite{roth1956form}. Moreover, attending to these patterns is also crucial as these patterns can be distributed across diverse locations in an epoch. For example, a particular waveform can be positioned at different locations, such as 1 s or 15 s on a 30-s signal. However, existing methods \cite{supratak2020tinysleepnet, phan2019seqsleepnet} mostly rely on the final output of single-scale convolutional layers to learn temporal representations, and consequently, they cannot fully capture the salient patterns \cite{wang2022novel}. In this study, we devise a novel deep learning structure that exploits a multi-scale representation and an attention mechanism for efficiently capturing salient waveforms.
	
	Furthermore, classifying transitioning epochs is also a challenging task in automatic sleep staging \cite{phan2019seqsleepnet, qu2020residual}. The difficult in classifying transitioning epochs arises from the fact that they usually contain mixed properties of multiple stages. For example, when a transition from N2 stage to N3 stage is in progress, such as the one shown in the second rectangle in Fig. \ref{transitioning_epochs}, it can contain the characteristics of both N2 and N3 stages, thereby confusing the classifier. To address this issue, previous studies \cite{ghimatgar2019automatic, phan2019seqsleepnet} exploited a hidden Markov model (HMM) or recurrent cell-based context encoder to implicitly model the inter-epoch relationships within learnable hidden states. Although the methods proposed in these studies proved effective in learning the latent context information, they still lacked the ability to efficiently identify confusing stages, struggling to classify transitioning epochs. Our proposed method not only exploits its inherent architectural strength but also two novel auxiliary tasks to explicitly learn the contextual representations, thereby efficiently identifying confusing stages. Specifically, we hypothesize that if the context encoder is aware of which stages are confusing and then exploits this information to adapt representations, it can classify stages more accurately. Therefore, we train our proposed network with an auxiliary stage classification task in an epoch-level manner. Additionally, we train our proposed network with the exact information regarding transitioning epochs through the other auxiliary task that detects stage transition. Thus, these two proposed auxiliary tasks allow our framework to infer information regarding which stages are confusing and whether transition occurred.
	
	The main contributions of the present study are summarized as follows:
	\begin{itemize}
		\item We propose a novel sleep staging network, TransSleep, to effectively capture intra-epoch	salient features as well as model inter-epoch contextual relationships.
		\item We exploit two auxiliary tasks, \emph{epoch-level stage classification} and \emph{stage-transition detection}, to address the problem of transitioning epochs during sleep
		staging.
		\item We conducted automatic sleep staging experiments on publicly available datasets
		and demonstrated the validity of our method by achieving state-of-the-art performances.
		\item We analyzed our proposed framework from diverse perspectives, thereby exploring the benefits of our proposed method.
	\end{itemize}
	
	\begin{figure*}[t]
		\centering
		\includegraphics[width=\linewidth]{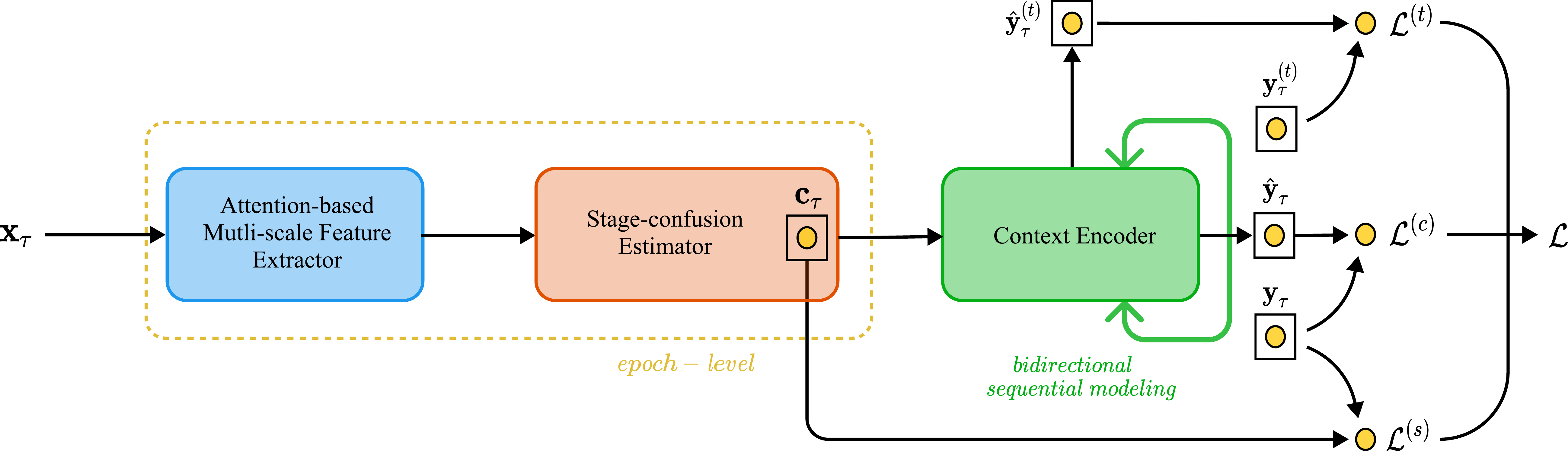}
		\caption{\label{network} Overall architecture of the proposed method, TransSleep. TransSleep consists of an epoch-level \emph{attention-based multi-scale feature extractor}, a \emph{stage-confusion estimator}, and a \emph{context encoder}. The proposed model exploits two auxiliary tasks to utilize their outputs to adapt feature representations, and therefore effectively discriminates confusing stages. In detail, class probabilities $\mathbf{c}_{\tau}$ in the stage-confusion estimator module infer which stages are confusing. Meanwhile, a predicted stage-transition vector $\hat{\mathbf{y}}_{\tau}^{(t)}$ in the context encoder provides explicit information about stage transition across neighboring epochs. The total objective function $\mathcal{L}$ consists of three loss functions: $\mathcal{L}^{(s)}$ for the epoch-level stage classification, $\mathcal{L}^{(t)}$ for the stage-transition detection, and $\mathcal{L}^{(c)}$ for the final sleep stage prediction $\hat{\mathbf{y}}_{\tau}$.}
	\end{figure*}
	
	This study is an extension of our recently accepted work for presentation at the 47\textsuperscript{th} IEEE International Conference on Acoustics, Speech, and Signal Processing\footnote{J. Phyo, W. Ko, E. Jeon, and H.-I. Suk, ``Enhancing Contextual Encoding with Stage-Confusion and Stage-Transition Estimation for EEG-based Sleep Staging,'' in \emph{Proc. 47th IEEE Int. Conf. Acoust. Speech Signal Process. (ICASSP)}, 2022, \url{https://2022.ieeeicassp.org/}.}. We supplement our original work by improving on the previous deep learning architecture for more effective salient waveform detection. We further exploit comparable baselines for a more substantial demonstration of the validity. We also visualize the loss curve of the stage-transition detection auxiliary task to show the training stability of the method. Finally, we conduct an ablation experiment and perform hypnogram analysis to evaluate our proposed method from diverse viewpoints.
	
	This remainder of this paper is organized as follows. Section \ref{realated work} discusses the previous studies on sleep stage classification. In Section  \ref{method}, we propose our novel deep learning-based sleep staging method involving two auxiliary tasks. Section \ref{experiments} describes the datasets used in this study and experimental settings in detail. We also report the quantitative results of classification in this section. In Section \ref{analysis}, we present qualitative and quantitative analyses of our framework. Finally, in Section \ref{conclusion}, we present our conclusions.
	
	\section{Related Work}
	\label{realated work}
	Representing stage-discriminative features of sleep EEG for automatic sleep staging remains challenging in both theory and practice. Pioneering studies have attempted to learn intra-epoch salient features as well as inter-epoch contextual features. In this section, we briefly discuss machine learning and deep learning-based methods for automatic sleep staging.
	
	\subsection{Machine Learning-based Autonomous Sleep Staging}
	Many studies have adopted traditional machine learning approaches for sleep stage classification. Tsinalis \textit{et al.} \cite{tsinalis2016automatic} manually extracted time-frequency features and exploited ensemble learning using stacked sparse autoencoders. Alickovic \textit{et al.} \cite{alickovic2018ensemble} performed discrete wavelet transformation for feature extraction and classified sleep stages using a modified SVM. Li \textit{et al.} \cite{li2017hyclasss} leveraged both intra- and inter-epoch features for sleep stage classification. The extracted time-frequency features were fed to a hybrid classifier, consisting of a random forest classifier and a Markov model. Ghimatgar \textit{et al.} \cite{ghimatgar2019automatic} selected optimal features based on the relevance and redundancy analysis and then classified them using a random forest classifier. Furthermore, an HMM was exploited to learn the transition between sleep stages.
	
	\subsection{Deep Learning-based Autonomous Sleep Staging}
	Recent advances made in deep learning have shown remarkable results in sleep staging. Multiple approaches to learn significant features from an EEG epoch have been proposed. Among many deep learning structures, a convolutional neural network (CNN)-based feature extractor has been widely used owing to its potential to represent spectro-temporal characteristics of EEGs. For instance, Supratak \textit{et al.} \cite{supratak2020tinysleepnet} employed single-scale CNNs to learn representative features of EEGs. Seo \textit{et al.} \cite{seo2020intra} adopted single-scale CNNs with residual connections to learn spectro-temporal properties of EEGs. Eldele \textit{et al.} \cite{eldele2021attention} applied two CNNs with small and large filter sizes on the first layer followed by single-scale CNNs. Qu \textit{et al.} \cite{qu2020residual} utilized two max-pooling layers of different sizes followed by single-scale CNNs. These approaches extracted distinctive features of EEGs using single-scale CNNs and only employed the final output without considering the multi-scale and locally distributed characteristics of the salient features of sleep EEGs.
	
	Some other previous studies \cite{supratak2017deepsleepnet, phan2019seqsleepnet} attempted to learn contextual dependencies from neighboring epochs during sleep staging. Inspired by sequence-to-sequence methods, these attempts achieved significant improvements owing to the inherent properties of sleep EEGs, such as transition rules \cite{berry2012rules, liang2012rule}. For instance, Supratak \textit{et al.} \cite{supratak2017deepsleepnet} proposed bidirectional long short-term memory (Bi-LSTM) to model contextual dependencies of neighboring epochs. Phan \textit{et al.} \cite{phan2019seqsleepnet} employed bidirectional gated recurrent unit (Bi-GRU) to encode contextual information on a sequence of extracted features. 
	%The Bi-GRU layer was used to model the temporal dependencies on the sequence of features. 
	Seo \textit{et al.} \cite{seo2020intra} extracted distinctive features in a subepoch-manner and employed Bi-LSTM to model contextual information. Their framework could learn intra- and inter-epoch contexts from raw single-channel EEGs from the temporal standpoint. Perslev \textit{et al.} \cite{perslev2019u} proposed a temporal fully convolutional network based on the U-Net \cite{ronneberger2015u} architecture. The network simultaneously represents stage-discriminative features as well as contextual properties. These existing methods utilized diverse context encoders, that is, Bi-LSTM and Bi-GRU, to embed contextual relationships of neighboring EEG epochs in learnable hidden states. Nevertheless, these methods did not directly focus on learning confusing stages and determining whether transitioning occurs explicitly.
	
	In this study, we propose a novel deep learning framework with two auxiliary tasks for effectively learning intra-epoch salient features as well as inter-epoch context properties. Furthermore, the proposed auxiliary tasks allow our deep learning architecture to determine which stages are confusing and whether a transition has occurred.
	
	\section{Method}
	\label{method}
	In this section, we propose a novel deep learning-based automatic sleep staging method, TransSleep, as schematized in Fig. \ref{network}. Our proposed framework consists of an attention-based multi-scale feature extractor (AMF) network, a stage-confusion estimator (SCE) network, and a context encoder (CE) network. The AMF extracts salient waveforms of input EEGs by adopting our proposed epoch-level temporal attention (ETA) module. Furthermore, two novel auxiliary tasks, an epoch-level stage classification task and a stage-transition detection task, are used for training the SCE and CE, respectively.
	
	\subsection{Problem Formulation}
	Automatic sleep staging is one of the most important issues in health informatics because human-centered (manual) sleep staging is an expensive procedure. Sleep EEG, a widely used measurement for sleep staging, has two crucial properties: (i) intra-epoch salient waveforms \cite{wang2022novel} and (ii) contextual relationships \cite{supratak2017deepsleepnet}. Hence, we propose a unified framework of a feature extractor and a context encoder from the viewpoint of sequence-to-sequence models.
	
	Let us denote the $\tau$\textsuperscript{th} sleep EEG epoch as $\mathbf{x}_\tau\in\mathbb{R}^T$ where $T$ is the number of time points in an epoch. Then, a feature extractor function represents an intra-epoch feature $\mathbf{f}_\tau\in\mathbb{R}^F$ from the input EEG, where $F$ denotes the dimension of the output feature. Our proposed AMF is designed to represent such intra-epoch salient patterns. Further, $\mathbf{f}_\tau$ is updated to $\tilde{\mathbf{f}}_\tau$ by SCE to improve modeling contextual properties. Subsequently, from the perspective of sequential modeling, a context encoder function encodes contextual relationships among neighboring features. Formally, the CE embeds a series of the represented features, $\mathbf{F}=[\mathbf{f}_\tau]_{\tau=1}^N=[\mathbf{f}_1,...,\mathbf{f}_N]$ and $\tilde{\mathbf{F}}=[\tilde{\mathbf{f}}_\tau]_{\tau=1}^N=[\tilde{\mathbf{f}}_1,...,\tilde{\mathbf{f}}_N]$ where $N$ denotes the number of epochs, and maps the embedding into the decision making layer to detect sleep stages $\hat{\mathbf{Y}}=[\hat{\mathbf{y}}_\tau\in\{0, 1\}^C]_{\tau=1}^N$, where $\hat{\mathbf{y}}_\tau$ is an one-hot encoded prediction of the $\tau$\textsuperscript{th} input EEG and $C$ is the number of sleep stages. In our proposed method, the SCE and CE
	model the contextual relationships among neighboring sequences. Furthermore, during the training of the SCE and CE, our proposed auxiliary tasks are used for providing explicit information about transitioning epochs.
	
	\subsection{Attention-based Multi-scale Feature Extractor}
	\subsubsection{Multi-scale Convolution Paths}
\begin{figure}[t]
		\centering
		\includegraphics[width=.875\columnwidth]{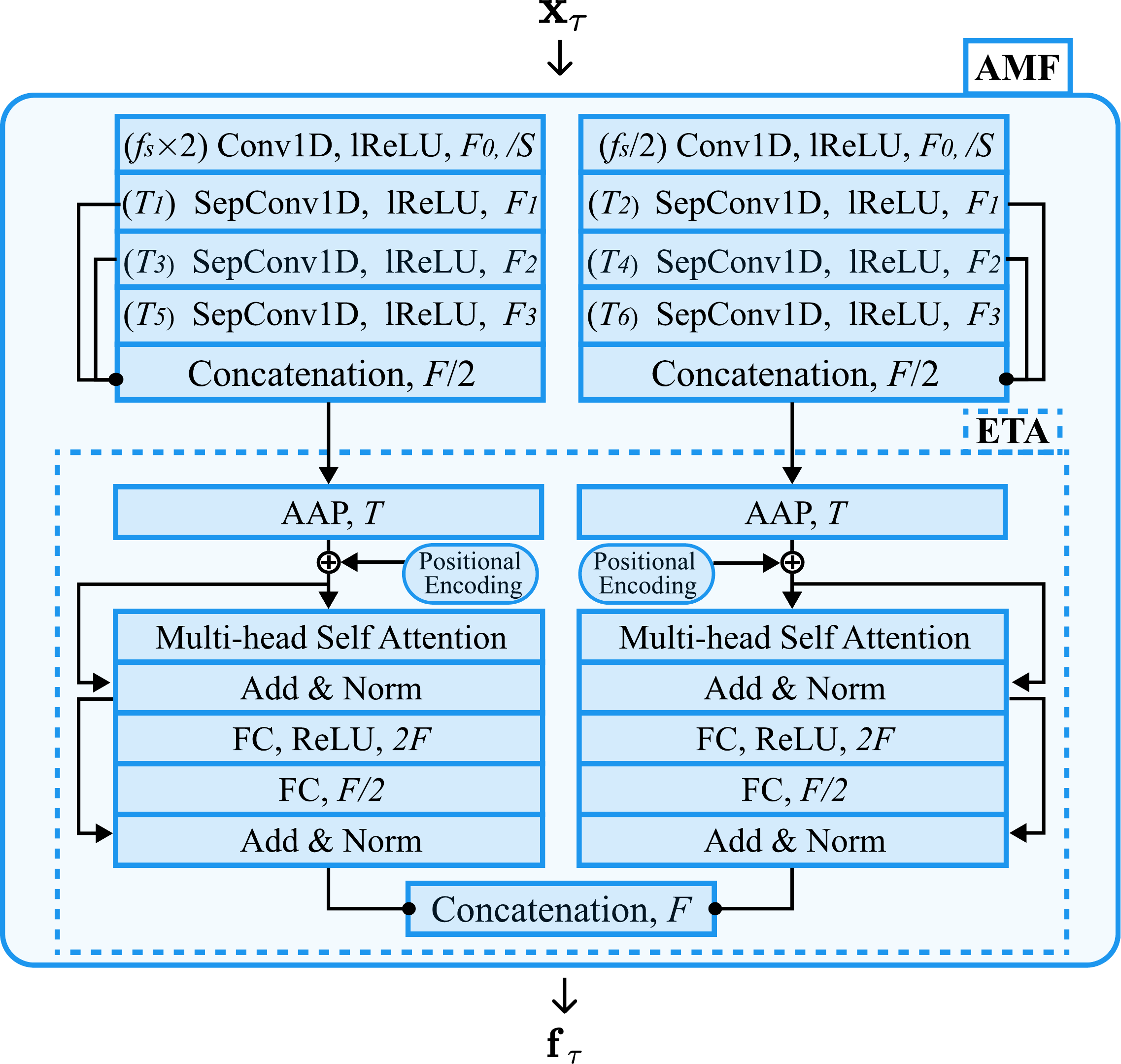}
		\caption{\label{amf} Design of the epoch-level feature extractor module. An input EEG $\mathbf{x}_\tau$ is fed into the feature extractor, and it outputs extracted feature $\mathbf{f}_\tau$. $f_s$ is a sampling frequency of a signal, $S$ is a stride, $T_k$ and $F_k$ are the kernel size and output feature maps of the $k$-th temporal separable convolution, $k$=1,2, $\cdots$ $l$, where $l$ is the number of temporal convolution layers. Each convolution block is followed by batch normalization, and AAP denotes adaptive average pooling.}
	\end{figure}

	To extract complex spectro-temporal features for each epoch, we design the AMF based on multi-scale CNNs \cite{ko2021multi} and a multi-head self-attention mechanism \cite{vaswani2017attention}, as depicted in Fig. \ref{amf}. As salient features have various spectro-temporal characteristics, we exploit a multi-scale structure by utilizing intermediate features for capturing multiple ranges of properties of input signals. Moreover, we set diverse kernel sizes for each convolution and use two feature-capturing paths, thereby allowing the AMF to recognize diverse salient patterns.
	
	Formally, each feature-capturing path uses a spectral convolution \cite{ko2021multi} and three separable temporal convolutions and outputs a $F/2$-dimensional output feature. Furthermore, we adopt two feature capturing paths with small and large filter sizes on the first spectral convolution layers, where each kernel can learn diverse frequency bands. This is motivated by previous studies that used different kernel sizes to learn different frequency properties \cite{jiang2018multiscale, supratak2017deepsleepnet}. Specifically, we use a convolution filter of size $(f_s\times2)$ and $(f_s/2)$ to capture low-frequency signals such as delta waves and high-frequency signals such as alpha waves due to the inverse proportional properties of time-frequency \cite{752051}, where $f_s$ denotes a sampling frequency of the input EEG signal. Then, to learn diverse multi-scale temporal trends of the input signals, we employ different kernel sizes for temporal convolutions.

	\subsubsection{Epoch-level Temporal Attention}
	After the multi-scale feature representation, our
	proposed AMF exploits an additional easy-to-adapt module, epoch-level temporal attention (ETA). The ETA module is proposed to highlight salient waveforms that appear at various points in an epoch. Therefore, our proposed method effectively represents salient features in an epoch-level manner.
	
	The ETA module uses adaptive average pooling  \cite{van2019evolutionary} on the features represented from the previous multi-scale paths. We then apply a multi-head self-attention mechanism with positional encoding \cite{vaswani2017attention} to focus on salient features. Finally, the attentive features are fed into two fully connected (FC) layers and the output features from two paths are concatenated, that is, $\mathbf{f}_\tau\in\mathbb{R}^F$.
	
	\subsection{Stage-Confusion Estimator}
	Existing studies mainly comprise two parts, extracting intra-epoch features and modeling contextual relationships among those features using learnable hidden representations \cite{supratak2017deepsleepnet, seo2020intra, ghimatgar2019automatic}. In this study, we propose the SCE to explicitly provide additional information about confusing stages before learning contextual relationships.
	
\begin{figure}[t]
		\centering
		\includegraphics[width=0.85\columnwidth]{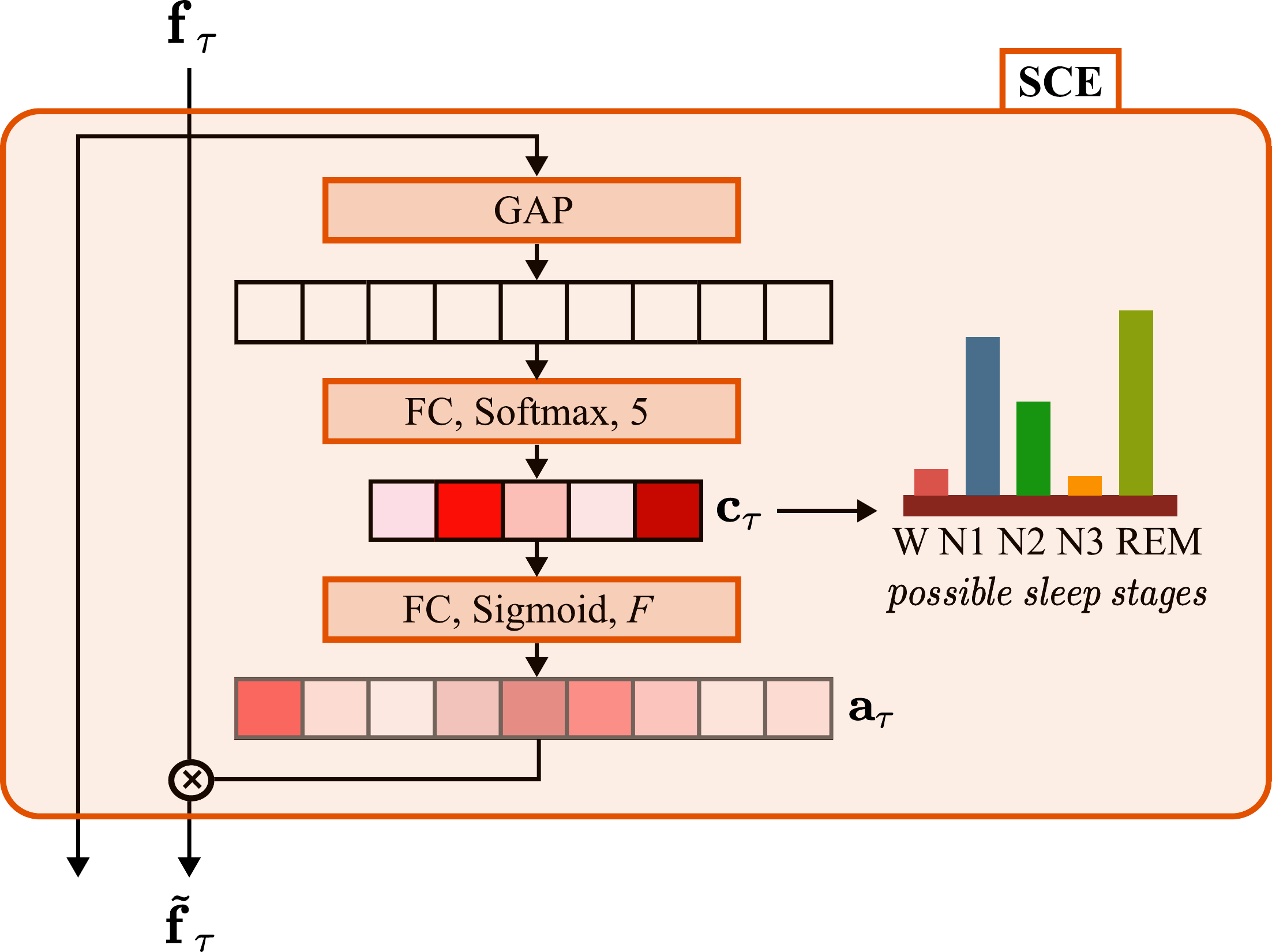}
		\caption{\label{sce} Design of the stage-confusion estimator module. Extracted feature $\mathbf{f}_\tau$ from the feature extractor is fed into the stage-confusion estimator and the updated feature vector $\tilde{\mathbf{f}}_\tau$ is obtained. GAP denotes global average pooling.}
	\end{figure}

	The SCE with an auxiliary task called epoch-level stage classification highlights the features related to the possible stages on each epoch via attention vectors. As illustrated in Fig. \ref{sce}, the SCE module estimates confusing stages through an auxiliary classifier in an epoch-level manner and reflects that information to the epoch-level feature representation via an attention mechanism.
	
	Particularly, given the epoch-level feature representation $\mathbf{f}_\tau$ estimated from the AMF, the SCE first computes a distribution of possible classes $\mathbf{c_\tau}$ using a logistic regression function as follows:
	\begin{equation}
		\mathbf{c}_{\tau}=\operatorname{softmax}(g(\operatorname{GAP}(\mathbf{f}_{\tau})))
	\end{equation}
	where $\operatorname{GAP}$ and $g$ denote the global average pooling \cite{lin2013network} and an FC layer used for the decision-making, respectively. 
	The individual class probability denotes the confidence of the respective class membership. Moreover, the class probabilities jointly carry the information regarding which classes are confusing. In this sense, we use the class probabilities $\mathbf{c}_\tau$ as a latent source of information to update the epoch-level feature representation with the attention mechanism. Specifically, we obtain an attention vector $\mathbf{a}_\tau$ from the class probabilities by applying a series of linear and non-linear operations as follows:
	\begin{equation}
		\mathbf{a}_\tau = \operatorname{sigmoid}(q(\mathbf{c}_\tau))
	\end{equation}
	where $q$ denotes an FC layer for exciting the confidence vector $\mathbf{c}_\tau$. Finally, the epoch-level feature representation is updated according to the attention vector via Hadamard multiplication, i.e., $\tilde{\mathbf{f}}_\tau = \mathbf{a}_\tau\otimes\mathbf{f}_\tau$. It is noteworthy that the attention-guided representation $\tilde{\mathbf{f}}_\tau$ entails both the signal-level features and the stage-confusion information.

	\subsection{Context Encoder}
	Sleep stages are labeled by considering specific transition patterns such as transition rules \cite{berry2012rules}. It is important to learn these patterns that appear over sleep. Accordingly, many conventional methods \cite{supratak2020tinysleepnet, qu2020residual} utilized RNN or self-attention to encode context information but without information of whether or not the transition occurred. Furthermore, other studies \cite{phan2018joint} attempted to predict stages of both contiguous epochs and a target epoch simultaneously by exploiting only the current epoch features. In contrast to these studies, we predict the final sleep stage by considering relations of neighboring epochs with the assistance of transition information obtained through auxiliary tasks.
	
	To enhance the modeling of inter-epoch contextual information, we encode temporal dependencies using the auxiliary task called stage-transition detection that provides explicit information about the transition. We design a CE module that embeds the inter-epoch relations from $[\tilde{\mathbf{f}}_\tau]_{\tau=1}^N$. Bi-LSTM cells \cite{schuster1997bidirectional} with an FC layer $r$ predict the stage-transition $\hat{\mathbf{y}}_\tau^{(t)}$ in a sequence-to-sequence manner:
	\begin{equation}
		[\hat{\mathbf{y}}_\tau^{(t)}]_{\tau=1}^N=r([\mathbf{h}_\tau]_{\tau=1}^N)
	\end{equation}
	where $[\mathbf{h}_\tau]_{\tau=1}^N=\operatorname{Bi-LSTM}([\tilde{\mathbf{f}}_\tau]_{\tau=1}^N)$.
	
	Basically, given a sequence of ground-truth stage labels in the training set, we can obtain the stage transition labels for the $\tau$-th epoch as follows:
	\begin{equation}
		\mathbf{y}_\tau^{(t)}=
		\begin{cases} 
			0 & \text{if } \mathbf{y}_{\tau-1} = \mathbf{y}_{\tau} = \mathbf{y}_{\tau+1}\\ 
			1 & \text{if } \mathbf{y}_{\tau-1} \neq \mathbf{y}_{\tau} \text{ or } \mathbf{y}_{\tau} \neq \mathbf{y}_{\tau+1}\\
		\end{cases}
	\end{equation}
	
	Armed with the additional information of stage-transition, we formulate the inter-epoch context encoder training in multitask learning. That is, the module is trained to optimally predict the sleep stages $\hat{\mathbf{y}}_\tau$ and the occurrence of transition $\hat{\mathbf{y}}_\tau^{(t)}$. Furthermore, our proposed CE also exploits originally represented feature $[\mathbf{f}_\tau]_{\tau=1}^N$ for richer representations.
	
	\begin{figure}[t]
		\centering
		\includegraphics[width=0.8\columnwidth]{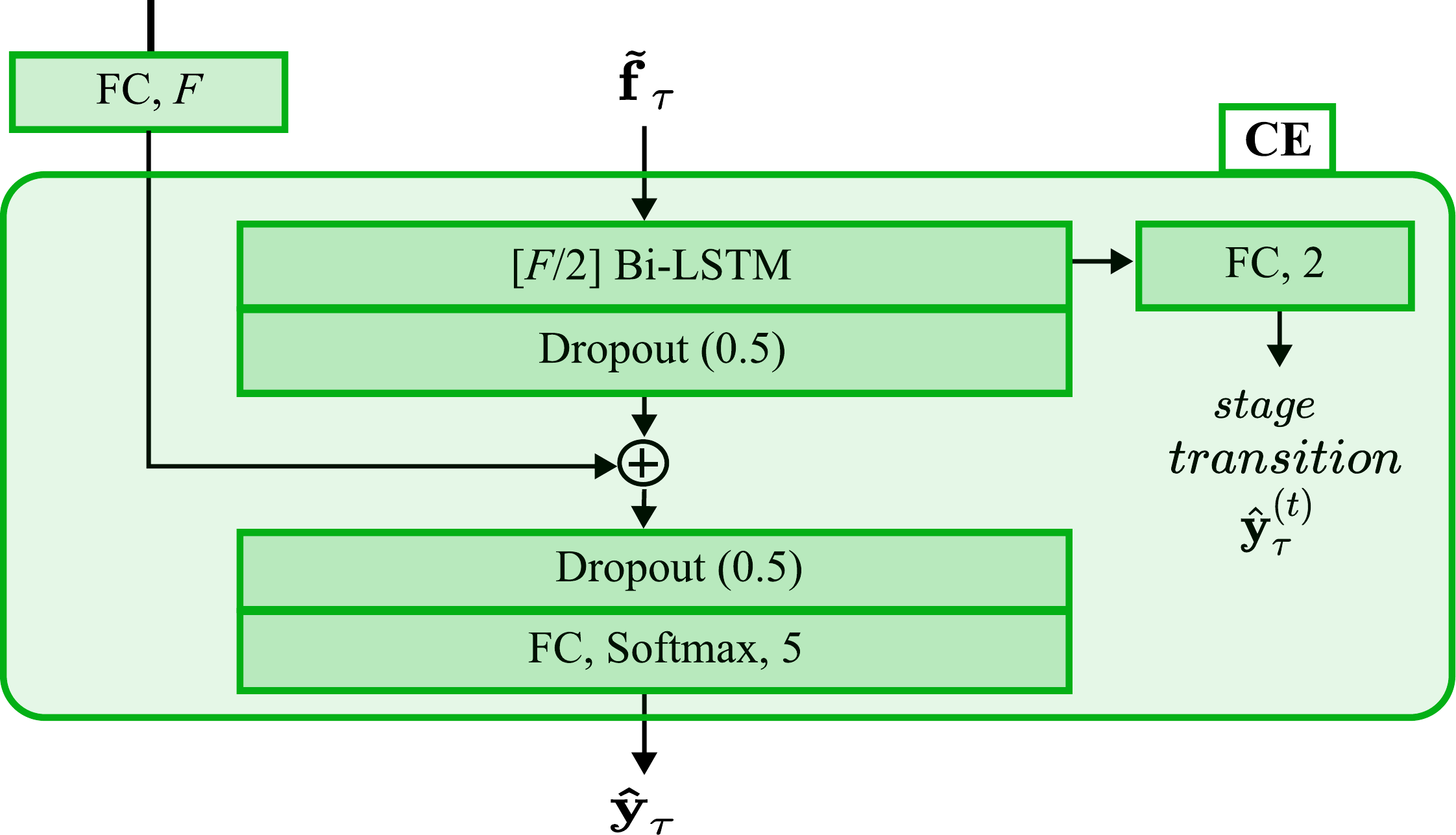}
		\caption{\label{ce} Design of the context encoder module. Updated feature $\tilde{\mathbf{f}}_\tau$ from the stage-confusion estimator is fed into the context encoder to obtain the predicted sleep stage $\hat{\mathbf{y}}_\tau$.}
	\end{figure}
	
	As schematized in Fig. \ref{ce}, we use a Bi-LSTM \cite{schuster1997bidirectional} architecture for the inter-epoch relations embedding and sequence-to-sequence classification. With regard to the transition detection and sleep stage classification, the hidden state $\mathbf{h}_\tau$ at each epoch is fed to the respective classifier to perform detections. Moreover, a residual connection is applied to add the input embedding to the output from Bi-LSTM. To regularize the CE module, two dropout layers with a rate of 0.5 are added after the Bi-LSTM and residual connection, respectively.
	
	\begin{algorithm}[t]
		\caption{\label{alg}Pseudo-algorithm for the proposed method}
		\KwIn{Traning dataset $\mathbb{D}=([\mathbf{x}_\tau]_{\tau=1}^{\cdots}, [\mathbf{y}]_{\tau=1}^{\cdots})$; intintial network parameters of the AMF, the SCE, and the CE $\theta_{\text{AMF}}$, $\theta_{\text{SCE}}$, $\theta_{\text{CE}}$; an optimization algorithm $\operatorname{Adam}$; coefficients $\lambda^{(c)}$, $\lambda^{(s)}$, and $\lambda^{(t)}$}
		\KwOut{Optimized parameters $\theta_{\text{AMF}}^\star$, $\theta_{\text{SCE}}^\star$, $\theta_{\text{CE}}^\star$}
		\While{network parameters not converged}{
			Draw a sequence $([\mathbf{x}_\tau]_{\tau=1}^N, [\mathbf{y}_\tau]_{\tau=1}^N)\sim\mathbb{D}$
			
			\For{$\tau=1,...,N$}{
				$\mathbf{f}_\tau\leftarrow\operatorname{AMF}(\mathbf{x}_\tau)$
				
				$\mathbf{c}_\tau\leftarrow\operatorname{softmax}(g(\operatorname{GAP}(\mathbf{f}_\tau)))${\ttfamily\ \#squeeze}
				
				$\mathbf{a}_\tau\leftarrow\operatorname{sigmoid}(q(\mathbf{c}_\tau))${\ttfamily\ \#excitation}
				
				$\mathcal{L}^{(s)}\leftarrow\operatorname{WCE}(\mathbf{y}_\tau, \mathbf{c}_\tau)$
				
				$\tilde{\mathbf{f}}_\tau\leftarrow\mathbf{a}_\tau\otimes\mathbf{f}_\tau$
				
				\If{$\mathbf{y}_{\tau-1}=\mathbf{y}_\tau=\mathbf{y}_{\tau+1}$}{
					$\mathbf{y}_\tau^{(t)}\leftarrow0$
				}
				\Else{$\mathbf{y}_\tau^{(t)}\leftarrow1$}
			}
			
			$[\mathbf{h}_\tau]_{\tau=1}^N=\operatorname{Bi-LSTM}([\tilde{\mathbf{f}}_\tau]_{\tau=1}^N)$
			
			$[\hat{\mathbf{y}}^{(t)}_\tau]_{\tau=1}^N\leftarrow r([\mathbf{h}_\tau]_{\tau=1}^N)$
			
			$\mathcal{L}^{(t)}\leftarrow\frac{1}{N}\sum\operatorname{WCE}(\mathbf{y}^{(t)}_\tau, \hat{\mathbf{y}}^{(t)}_\tau)$
			
			$[\hat{\mathbf{y}}_\tau]_{\tau=1}^N\leftarrow\operatorname{CE}([\mathbf{f}_\tau]_{\tau=1}^N, [\tilde{\mathbf{f}}_\tau]_{\tau=1}^N)$
			
			$\mathcal{L}^{(c)}\leftarrow\frac{1}{N}\sum\operatorname{WCE}(\mathbf{y}_\tau, \hat{\mathbf{y}}_\tau) + \lambda^{(c)}\cdot\operatorname{WCS}(\mathbf{y}_\tau, \hat{\mathbf{y}}_\tau)$
			
			$\mathcal{L}\leftarrow\mathcal{L}^{(c)}  + \lambda^{(s)}\mathcal{L}^{(s)} + \lambda^{(t)}\mathcal{L}^{(t)}$
			
			$\Theta\leftarrow\theta_{\text{AMF}}\cup\theta_{\text{SCE}}\cup\theta_{\text{CE}}$
			
			$\Theta\leftarrow\Theta+\operatorname{Adam}(\mathcal{L})$
		}
	\end{algorithm}

	\subsection{Optimization}
	To optimize the tunable parameters of AMF, SCE, and CE, we jointly exploit three learning tasks: the main downstream task and two auxiliary tasks that provide explicit information about transitioning epochs, that is, the epoch-level stage classification and the stage-transition detection tasks.
	
	\subsubsection{Epoch-level Stage Classification}
	The epoch-level stage classification task is defined as a multi-class classification for sleep stage similar to the main downstream task. Note that this auxiliary task is different from the main task, albeit its similarity, because this auxiliary task is only performed on a single epoch whereas the main task calculates the loss value of sequences.
	
	Here, the objective function $\mathcal{L}^{(s)}$ for this auxiliary task is estimated using a class-weighted cross-entropy ($\operatorname{WCE}$) function:
	\begin{equation}
		\mathcal{L}^{(s)} = \operatorname{WCE}(\mathbf{y}_\tau, \mathbf{c}_\tau) = -w_c\sum_{\mathbf{c}_\tau\in\mathcal{C}}\mathbf{y}_\tau\cdot\log{{\mathbf{c}_\tau}}
	\end{equation}
	where $\mathbf{y}_\tau$ and $\mathbf{c}_\tau$ denote the ground truth and the predicted label estimated by the SCE for the $\tau$\textsuperscript{th} input EEG, $w_c$ is a scaling factor that is the inverse proportion of $c$-class samples in the training set, and $\mathcal{C}=\{1, 2, ..., C\}$ is a set of possible classes. This task provides explicit information about which stages are confusing for an epoch to our proposed framework and thus improves the ability of the framework to identify confusing stages.
	
	\subsubsection{Stage-transition Detection}
	The stage-transition detection task is defined as a binary classification for stage transition. The objective function of this auxiliary task is calculated using a WCE function:
	\begin{equation}
		\mathcal{L}^{(t)} = \frac{1}{N}\sum_{\tau=1}^{N}\operatorname{WCE}(\mathbf{y}^{(t)}_\tau, \hat{\mathbf{y}}^{(t)}_\tau)
	\end{equation}
	where $\mathbf{y}^{(t)}$ is an auxiliary stage-transition label and $\hat{\mathbf{y}}^{(t)}$ is a predicted label estimated from hidden representation of Bi-LSTM. As this auxiliary task directly focuses on detecting stage-transition, our proposed framework can be well aware of contextual relationships among transitioning epochs.
	
	\subsubsection{Downstream Task}
	The main downstream task involves multi-class classification. To attenuate the class-imbalance issue, we also use a WCE function and a class-weighted cosine similarity (WCS) objective function \cite{nguyen2010cosine}. The objective function for the downstream task is expressed as follows:
	\begin{align}
		\mathcal{L}^{(c)}=\frac{1}{N}\sum_{\tau=1}^N&\operatorname{WCE}(\mathbf{y}_\tau, \hat{\mathbf{y}}_\tau) + \lambda^{(c)}\cdot\operatorname{WCS}(\mathbf{y}_\tau, \hat{\mathbf{y}}_\tau)\nonumber,
		\\
		\operatorname{WCS}(\mathbf{y}_\tau, \hat{\mathbf{y}}_\tau)&=-w_c\sum_{\mathbf{c}_\tau\in\mathcal{C}}(1-\cos(\mathbf{y}_\tau, \hat{\mathbf{y}}_\tau))
	\end{align} 
	where $\cos(\mathbf{v}, \mathbf{w})=\mathbf{v}\cdot\mathbf{w}/\|\mathbf{v}\|\cdot\|\mathbf{w}\|$ is the cosine similarity operation and $\lambda^{(c)}$ denotes a scaling hyperparameter.

	Owing to the two auxiliary tasks, our proposed framework can learn richer representations for the main sleep staging task. The final objective function is expressed as follows:
	\begin{equation}
		\label{total_loss}
		\mathcal{L} = \mathcal{L}^{(c)}  + \lambda^{(s)}\mathcal{L}^{(s)} + \lambda^{(t)}\mathcal{L}^{(t)} 
	\end{equation}
	where $\lambda^{(s)}$ and $\lambda^{(t)}$ are coefficient constants to control balance of each loss term. The complete pseudo-algorithm to train all the networks of the proposed framework is presented in Algorithm \ref{alg}.
	
	\begin{table}[t]
		\centering
		\caption{\label{dataset}Statistics of Sleep-EDF and MASS datasets}
		\vspace{.1cm}
		\begin{tabular}{|c|c|c|}
			\hline 
			Dataset & Sleep-EDF & MASS \\
			\hline 
			Subjects & 20 & 62\\
			Channel & Fpz-Cz & F4-LER\\
			Sampling rate & $100 \mathrm{~Hz}$ & $256 \mathrm{~Hz}$\\
			\hline
			W & 8285 (19.6 \%) & 6231 (10.6 \%)\\
			N1 & 2804 (6.6 \%) & 4814 (8.2 \%)\\
			N2 & 17799 (42.1 \%) & 29777 (50.4 \%)\\
			N3 & 5703 (13.5 \%) & 7653 (12.9 \%)\\
			REM & 7717 (18.2 \%) & 10581 (17.9 \%)\\\hline
			Total & 42308 & 59056 \\
			\hline
		\end{tabular}
	\end{table}

	\begin{table*}[t]
		\centering
		\caption{\label{performance}Performance comparison between TransSleep and state-of-the art methods.}
		\vspace{.1cm}
		\begin{tabular}{|c|c|c|cc|ccccc|}
			\hline 	
			\multirow{2}{*}{\shortstack{\textbf{Dataset}}} &
			\multirow{2}{*}{\shortstack{\textbf{Channel}}} &
			\multirow{2}{*}{\shortstack{\textbf{Method}}} &
			\multicolumn{2}{|c|}{\textbf{Overall Metrics}} & \multicolumn{5}{c|}{\textbf{Per-class F1}} \\
			&  &  & ACC & MF1 & W & N1 & N2 & N3 & REM \\
			
			\hhline{|=|=|=|==|=====|} 
			\multirow{9}{*}{\shortstack{Sleep-EDF}} &
			\multirow{9}{*}{\shortstack{Fpz-Cz}} &
			SAE \cite{tsinalis2016automatic} & 
			78.9 & 73.7 & 71.6 & 47.0 & 84.6 & 84.0 & 81.4 \\
			& & DeepSleepNet \cite{supratak2017deepsleepnet} & 
			82.0 & 76.9 & 86.0 & 45.0 & 85.1 & 84.0 & 82.6 \\
			& & SeqSleepNet \cite{phan2019seqsleepnet}  & 85.2 & 78.4 & - & - & - & - & -   \\
			& & JointCNN \cite{phan2019seqsleepnet} & 81.9 & 73.8 & - & - & - & - & -   \\
			& & IITNet \cite{seo2020intra} & 83.9 & 77.6 & 87.7 & 43.4 & 87.7 & 86.7 & 82.5\\
			& & U-time \cite{perslev2019u} & - & 78.6 & 87.1 & 51.5 & 86.4 & 84.2 & 83.7 \\
			& & ResnetMHA \cite{qu2020residual} & 84.3 & 79.0 & \textbf{90.2} & 48.3 & \bf87.8 & 85.6 & 83.0  \\
			%		& & Eldele \cite{eldele2021attention} & \bf 85.6 & 80.9 & 90.3 & 47.9 & 89.8 & 89.0 & 85.0  \\
			& & XSleepNet \cite{phan2021xsleepnet} & \bf 86.3 & 80.6 & - & - & - & - & -  \\\cline{3-10}
			& & \textit{Proposed} & 86.1 & \textbf{81.7} & 90.0 & \textbf{55.5} & \bf 87.8 & \bf88.7 & \textbf{86.3}  \\	
			
			\hhline{|=|=|=|==|=====|}
			\multirow{5}{*}{\shortstack{MASS}} &
			\multirow{5}{*}{\shortstack{F4-LER}} &
			DeepSleepNet \cite{supratak2017deepsleepnet} & 86.7 & 81.2 & 87.5 & 55.4 & 91.3 & 84.8 & \bf 87.2 \\ 
			& & SeqSleepNet \cite{phan2019seqsleepnet} & 82.8 & 77.6 & 82.9 & 54.6 & 87.1 & 79.3 & 83.8 \\
			& & U-time \cite{perslev2019u} & 85.6 & 80.5 & 85.4 & 58.6 & 89.9 & 83.9 & 87.0\\
			& & ResnetMHA \cite{qu2020residual} & 86.5 & 81.0 & 87.2 & 52.9 & \bf 91.5 & \bf 87.0 & 86.6  \\\cline{3-10}
			& & \textit{Proposed} & \bf 87.4 & \bf 82.6 & \bf 87.9 & \textbf{61.3} & 91.3 & 85.3 & \bf 87.2  \\
			\hline
		\end{tabular}
	\end{table*}

	\section{Experiments}
	\label{experiments}
	In this section, we describe the sleep EEG datasets used in our experiments. We further report detailed experimental settings and the performance of our method in comparison with competing methods. Our experimental code is available at: \url{https://github.com/ku-milab/TransSleep}.

	\subsection{Datasets}
	We used two publicly available EEG datasets for sleep staging to demonstrate the validity of our proposed method.
	
	\subsubsection{Sleep-EDF} Sleep-EDF \cite{goldberger2000physiobank}, a widely used dataset for sleep analysis, contains whole-night PSG recordings acquired from Fpz-Cz and Pz-Oz EEG channels and horizontal EOG channel at a sampling rate of 100 Hz, and EMG at a sampling rate of 1 Hz. Similar to previous studies \cite{supratak2017deepsleepnet, perslev2019u}, we used a single-channel EEG of Fpz-Cz recorded from 20 healthy subjects, 25 to 34 years old. These recordings were manually labeled into one of eight classes (W, N1, N2, N3, N4, REM, MOVEMENT, and UNKNOWN) by sleep experts. The N3 and N4 stages were merged into N3 to ensure consistency with MASS \cite{o2014montreal}, and the epochs annotated as MOVEMENT or UNKNOWN were discarded; therefore, $C=5$. In addition, as there were long periods of stage W at the start and end of each recording, continuous wake epochs longer than $30$ min outside the sleep period were ignored in our experiments, as is the case with competing baselines \cite{supratak2017deepsleepnet, perslev2019u}.
	
	\subsubsection{MASS} We further conducted sleep staging experiments on the third subsection of the MASS \cite{o2014montreal} dataset. This dataset contains PSG recordings from 62 healthy subjects, 20 to 69 years old. Each recording is acquired from 20 EEG electrodes referenced by A2 electrode, 2 EOG electrodes, and 3 EMG electrodes. Similar to existing studies \cite{supratak2017deepsleepnet, qu2020residual}, we used a single F4-LER channel for fair comparisons. The signals in this dataset were recorded at a sampling rate of either 256 Hz or 512 Hz. We downsampled all signals to 100 Hz.
	
	We preprocessed the Sleep-EDF and MASS signals by performing quantile normalization (median: 0 and inter quantile range: 1) and band-pass filtering within 0.5 to 49 Hz. TABLE \ref{dataset} shows the statistics of the two datasets.
	
	\subsection{Experimental Settings}
	\subsubsection{Proposed Method}
	We evaluated our proposed method using a $k$-fold cross-validation scheme, where $k$ was set to 20 and 31 for the Sleep-EDF and MASS datasets, respectively. Similar to competing baselines \cite{perslev2019u, phan2018joint}, we used \emph{leave-one-subject-out cross-validation} and \emph{leave-two-subjects-out cross-validation} for the Sleep-EDF and MASS datasets, respectively. 
	
	For the AMF, the stride size was set to $2$, that is, $S=2$. We set the kernel size as $T_1=11$, $T_2=15$, $T_3=7$, $T_4=9$, $T_5=3$, and $T_6=5$, whereas the dimensions of the output feature $F_0=4$, $F_1=16$, $F_2=32$, and $F_3=64$. Note that the output feature size of each convolution path is $F=2\times\sum_{i=1}^{3}F_i=224$. Thus, the number of hidden nodes in the SCE was also $224$. Finally, we calculated the contextual information of $N=25$ epochs; hence, the number of Bi-LSTM cells was $25$.
	
	We employed an Adam optimizer \cite{kingma2014adam} with a learning rate of $10^{-3}$ for the loss function defined in Eq. (\ref{total_loss}). We set $\beta_1$ as 0.9 and $\beta_2$ as 0.999. We also regularized all tunable parameters using a Ridge regularizer ($\ell_2=0.001$) \cite{hoerl1970ridge}. We controlled the objective functions via $\lambda^{(c)}=2$, $\lambda^{(s)}=2$ and $\lambda^{(t)}=0.2$. We set the batch size and the number of training iterations as 32 and 150, respectively. We applied the early stopping strategy based on the validation F1 performance.
	
	\subsubsection{Baselines}
	\label{baseline}
	In our experiments, we compared our proposed method with eight state-of-the-art baselines, a stacked sparse autoencoder (SAE; \cite{tsinalis2016automatic}), DeepSleepNet \cite{supratak2017deepsleepnet}, SeqSleepNet \cite{phan2019seqsleepnet}, JointCNN \cite{phan2018joint}, IITNet \cite{seo2020intra}, U-time \cite{perslev2019u}, ResnetMHA \cite{qu2020residual}, and XSleepNet \cite{phan2021xsleepnet}
	
	\subsection{Experimental Results}
	\subsubsection{Evaluation Metrics}
	To evaluate the performance of the model for sleep stage classification, we adopted two metrics, namely, the macro-averaged F1-score (MF1) and accuracy (ACC). We further computed per-class metrics by considering a single class as a positive class and all other classes combined as a negative class. MF1 and ACC can be calculated as follows:
	\begin{equation}
		\mathrm{MF}1=\frac{\sum_{c=1}^{C} \mathrm{~F} 1_{c}}{C}\text{\quad,\quad}
		\mathrm{ACC}=\frac{\sum_{c=1}^{C} \mathrm{TP}_{c}}{M}
	\end{equation}
	where $\mathrm{~F} 1_{c}$ is per-class F1-score of class $c$, $C$ is the number of sleep stages, $\mathrm{TP}_{c}$ is the true positives of class $c$, and $M$ is the total number of EEG epochs. 
	
	\begin{figure*}[h]
		\centering
		\includegraphics[width=\linewidth]{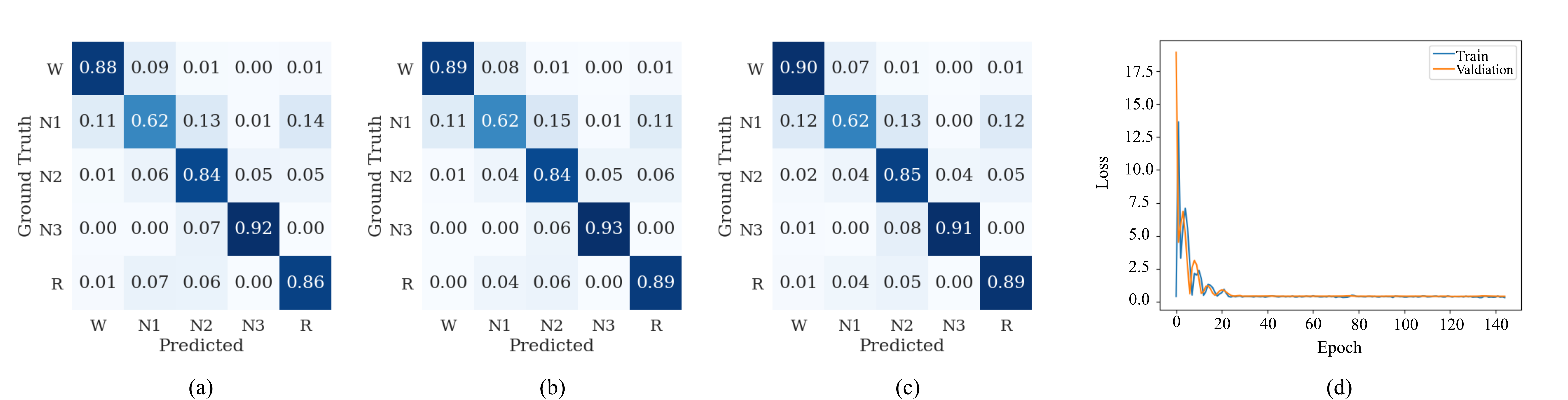}
		\caption{\label{confusion_loss_curve} Confusion matrix from 20-fold cross-validation on the Sleep-EDF dataset: (a) Case I, (b) Case II, (c) TransSleep in the ablation experiments, and (d) loss curves for the stage-transition detection auxiliary task.}
	\end{figure*}
	
	\begin{figure}[t]
		\centering
		\includegraphics[width=\columnwidth]{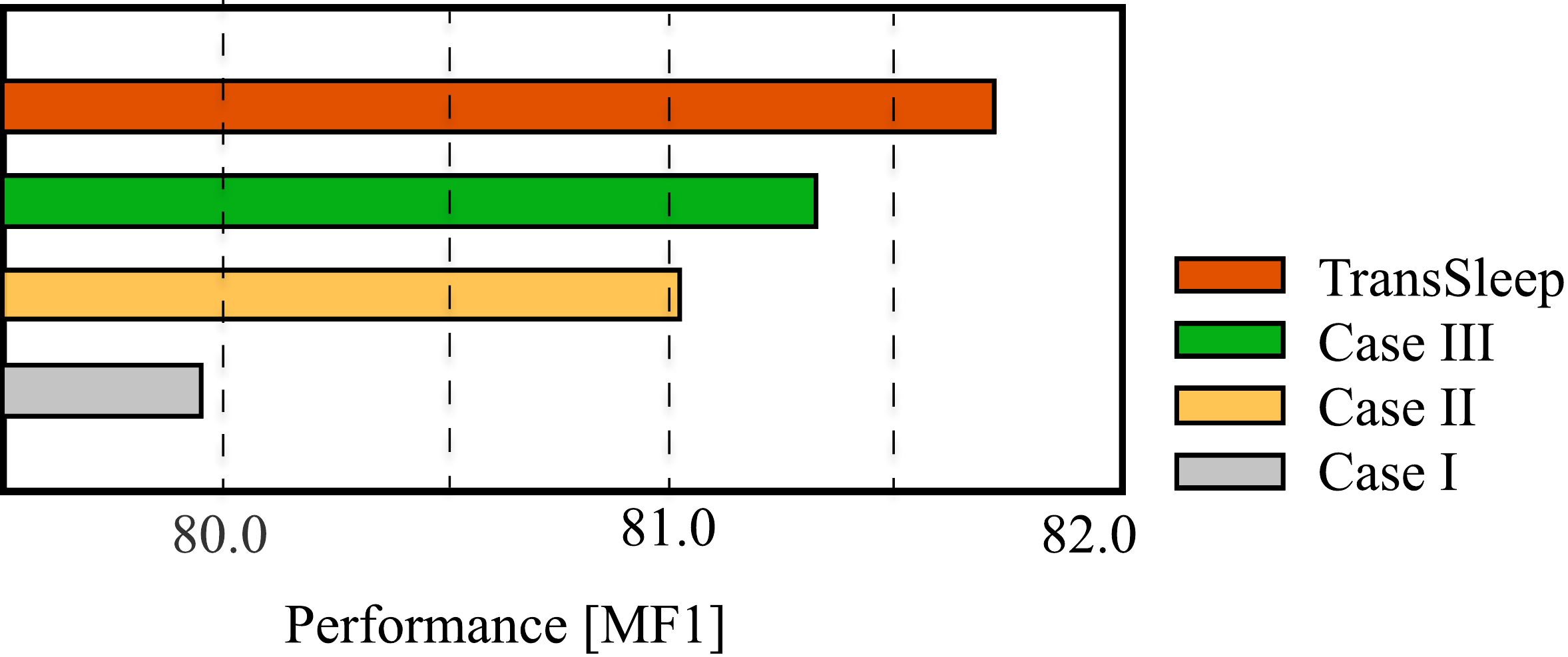}
		\caption{\label{ablation} Macro-averaged F1 scores of ablation cases on the Sleep-EDF dataset: Case I (gray), Case II (yellow), Case III (green), and our proposed TransSleep (red).}
	\end{figure}

	\subsubsection{Results}
	We compared our proposed TransSleep with competing baseline methods on the Sleep-EDF and MASS datasets, as summarized in TABLE \ref{performance}. Our proposed TransSleep outperformed competing state-of-the-art baselines except for some cases. Specifically, for the Sleep-EDF dataset, our proposed method achieved the best MF1 and four per-class F1 scores. Moreover, our proposed method showed the second best score for the per-class F1 of W stage and ACC scores. Additionally, our proposed method also showed the best performance for ACC, MF1, and three per-class F1 scores for the MASS dataset.
	
	Note that the N1 and REM stages are almost indistinguishable by visual inspection \cite{corsi2006power, an2021effective}. For these confusing stages, our proposed method obtained promising improvements compared to other baselines. It is worth noting that N1 is the most difficult sleep stage to classify because N1 commonly occupies the lowest proportion during sleep. Thus, the highest per-class F1 scores of N1 stage confirm that our proposed method was not biased to the majority of sleep stages. Moreover, from the practical standpoint, REM is an essential stage for diagnosing several sleep disorders \cite{gagnon2002rem, st2017rem}. Hence, we expect that our proposed method is clinically meaningful for practical diagnosis and assessments of sleep disorders.
	
	Based on these promising results, we concluded that our proposed architectural design and two auxiliary tasks played vital roles in identifying confusing stages by enhancing the representations of intra-epoch salient features and inter-epoch contextual relationships.
	
	\section{Analysis}
	\label{analysis}
	We analyzed our proposed method from diverse perspectives. We conducted ablation experiments to explore the validity of our proposed architectural design and auxiliary tasks. We also examined the ability to identify transitioning epochs in both qualitative and quantitative manners by observing hypnograms and estimating the proportions of misclassified epochs in transitioning and non-transitioning epochs.
	
	\begin{figure*}[t]
		\centering
		\includegraphics[width=.85\linewidth]{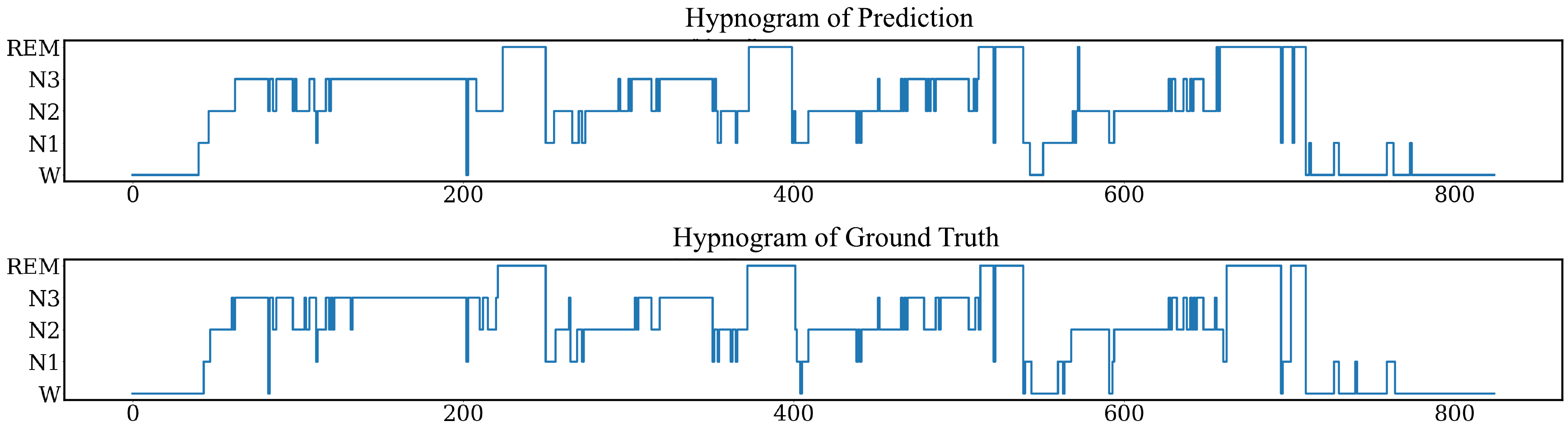}\\
		{\small (a) Hypnogram produced by our proposed TransSleep (top; ACC: $89.7$ and MF1: $84.5$) and the ground truth (bottom).}
		\vskip.15in	
		\includegraphics[width=.85\linewidth]{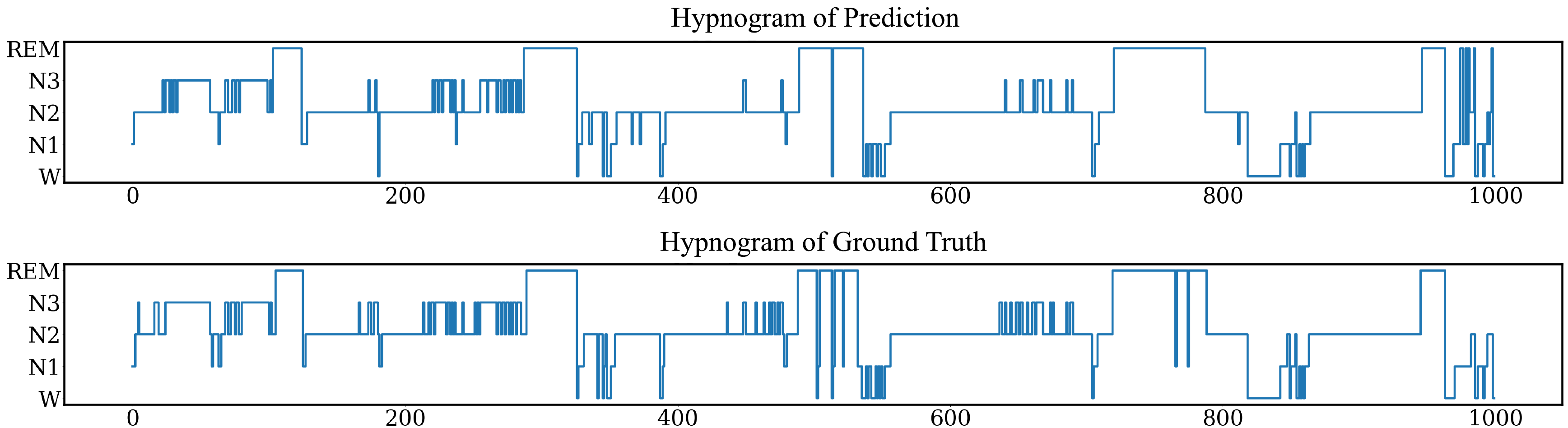}\\
		{\small (b) Hypnogram produced by our proposed TransSleep (top; ACC: $89.9$ and MF1: $85.8$) and the ground truth (bottom).}
		\vskip.15in		
		\caption{\label{hypnogram}Hypnograms of the first subject from (a) Sleep-EDF dataset and (b) MASS dataset.}
	\end{figure*}
	
	\subsection{Ablation Experiments}
	In this study, we proposed a novel feature extractor component, namely, the ETA in AMF, for intra-epoch salient feature representation, and two auxiliary tasks, namely, the epoch-level stage classification and the stage-transition detection, to correctly detect confusing stages. To verify the effects of these components, we conducted ablation experiments with three variants:
	\begin{itemize}
		\item Case I: We constructed a vanilla TransSleep without the ETA, and hence, the AMF is composed of only multi-scale CNNs. Furthermore, we removed two auxiliary tasks; therefore, $\mathcal{L} = \mathcal{L}^{(c)}$.
		\item Case II: For this case, we exploited the ETA structure but still did not use the two auxiliary tasks; therefore, $\mathcal{L} = \mathcal{L}^{(c)}$.
		\item Case III: We only removed the stage-transition detection task; therefore, $\mathcal{L} = \mathcal{L}^{(c)} + \lambda^{(s)}\mathcal{L}^{(s)}$.
		\item TransSleep: We used our proposed method; therefore, $\mathcal{L} = \mathcal{L}^{(c)} + \lambda^{(s)}\mathcal{L}^{(s)} + \lambda^{(t)}\mathcal{L}^{(t)}$.
	\end{itemize}
	
	As indicated in Fig. \ref{ablation}, the proposed ETA module and two auxiliary tasks are beneficial for sleep staging improvements. By comparing Case I and Case II, we concluded that the proposed ETA improves the intra-epoch salient feature representation as it captures locally distributed distinctive waveforms by attending to specific parts of a signal.
	
	Additionally, each auxiliary task demonstrated its efficiency in sleep staging. We observed the ablation of the epoch-level stage classification task by comparing Case II and Case III. Here, information about confusing stages estimated by the SCE clearly improves the performance of our proposed method by informing about the confusing stages, thereby modeling richer contextual relationships among neighboring epochs.
	
	We further examined the ability of the stage-transition detection task by comparing Case III and TransSleep. The results showed that detecting stage transition helps our proposed method represent contextual information effectively.
	
	Our overall findings reveal that TransSleep, comprising an architecturally easy-to-adapt ETA module and two plug-in auxiliary tasks, significantly improves the representations of salient patterns and contextual relationships. TransSleep, in fact, outperformed all other ablations.
	
	For further analysis, we calculated confusion matrices obtained from 20-fold cross-validation results on the Sleep-EDF dataset. Specifically, we reported Case 1 (the vanilla model without the ETA and two auxiliary tasks), Case II (our model without auxiliary tasks), and the proposed method in Fig. \ref{confusion_loss_curve}. The rows and columns show the ground truth and predicted class, respectively. We observed that the confusion matrix of our proposed method shows a relative increase in correctly classified epochs and a relative decrease in misclassified epochs of contiguous stages in transitioning phases compared to the others.
	
	Further, we plotted loss curves of the stage-transition detection task in Fig. \ref{confusion_loss_curve} to identify whether the context encoder actually learns the transition information. We observed that both training and validation loss curves converge, which indicates that the proposed model efficiently detects stage transitions.

	\subsection{Hypnogram}
	We plotted a hypnogram of the predicted stages and the ground truth stages in Fig. \ref{hypnogram}, which shows the stages of sleep as a function of time. We observed that hypnograms of TransSleep are similar to the ground truths for both the Sleep-EDF and MASS datasets. This implies that the proposed method efficiently classifies sleep stages.
	
	For quantitative analysis, we further calculated the proportion of misclassified samples in transitioning epochs on the Sleep-EDF dataset, as reported in TABLE \ref{te}. We defined transitioning epochs as epochs in which stages are different from adjacent epochs. On the Sleep-EDF dataset, transitioning samples and non-transitioning samples exhibit ratios of 17.3\% and 83.7\%, respectively. In this experiment, TransSleep achieved lower misclassification errors for both transitioning and non-transitioning epochs compared with Case II. Particularly, in transitioning epochs, TransSleep misclassified 35.7\% and Case II misclassified 37.8\%, whereas TransSleep misclassified 10.4\% and Case II misclassified 11.4\% in non-transitioning epochs. Note that detecting transitioning epochs is more difficult than non-transitioning ones for both Case II and TransSleep. Despite the difficulty in classifying transitioning epochs, our proposed method outperformed Case II by 2.1\% in transitioning epochs, while the difference was 0.9\% in non-transitioning epochs. Thus, the results showed that TransSleep well classifies transitioning epochs.
	
		\begin{table}[t]
		\centering
		\caption{\label{te} Proportion of misclassified epochs of the first subject in transition/non-transition epochs for the Sleep-EDF dataset.}
		\vspace{.1cm}
% 		\resizebox{\columnwidth}{!}{%
\begin{adjustbox}{width=\columnwidth,center}
        \begin{tabular}{|c|cc|c|}
			\hline 
			\multirow{2}{*}{\shortstack{\textbf{Types}}} & \multicolumn{2}{c|}{\textbf{Variants}} & \multirow{2}{*}{\shortstack{\textbf{Epochs }}}  \\
			& Case II & TransSleep &\\
			\hline 
			Transitioning & 37.8 (\%) & 35.7 (\%) & 17.3 (\%)\\
			\hline
			Non-transitioning  & 11.3 (\%) & 10.4 (\%) & 83.7 (\%)\\
			\hline
		\end{tabular}%
		\end{adjustbox}
% 		}
	\end{table}
	\section{Conclusion}
	\label{conclusion}
	We proposed the TransSleep network, which includes two auxiliary tasks, for automatic sleep staging. TransSleep can effectively capture salient waveforms by focusing on mining distinct local patterns in sleep EEG. Moreover, two auxiliary tasks, epoch-level stage classification and stage-transition detection, improve the ability of TransSleep to discriminate confusing stages. We conducted sleep staging experiments on two publicly available datasets, thereby demonstrating the validity of our proposed method. Furthermore, we analyzed TransSleep from different perspectives, and our results show that TransSleep can be helpful in providing new insights into deep learning-based sleep staging.
	
	TransSleep, albeit its potential, suffers from a few limitations. We expect that a layer-wise hierarchical classifier can classify confusing stages, and from a health informatics viewpoint, interpreting the decision-making process is critical. We intend to explore these aspects in our forthcoming research.

\begin{acknowledgment}
    This work was supported by Institute for Information \& Communications Technology Promotion (IITP) grant funded by the Korea government under Grant 2017-0-00451 (Development of BCI based Brain and Cognitive Computing Technology for Recognizing User’s Intentions using Deep Learning) and Grant 2019-0-00079 (Department of Artificial Intelligence, Korea University).
\end{acknowledgment}

% \bibliographystyle{arx}
% \bibliography{BIBarx}
\bibliography{arx}
\bibliographystyle{ieeetr}
\end{document}